\documentclass[letterpaper]{article} 
\usepackage{aaai25}  
\usepackage{times}  
\usepackage{helvet}  
\usepackage{courier}  
\usepackage[hyphens]{url}  
\usepackage{graphicx} 
\usepackage{natbib}  
\usepackage{caption} 
\usepackage{algorithm}
\usepackage{algorithmic}

\usepackage{newfloat}
\usepackage{listings}
\DeclareCaptionStyle{ruled}{labelfont=normalfont,labelsep=colon,strut=off} 
\floatstyle{ruled}
\newfloat{listing}{tb}{lst}{}
\floatname{listing}{Listing}
\usepackage{algorithm}
\usepackage{algorithmic}
\usepackage{amsmath,amsfonts}
\usepackage{adjustbox} 
\usepackage{algorithmic}
\usepackage{array}
\usepackage[caption=false,font=normalsize,labelfont=sf,textfont=sf]{subfig}
\usepackage{textcomp}
\usepackage{url}
\usepackage{verbatim}
\usepackage{graphicx}
\usepackage{textcomp} 
\usepackage[T1]{fontenc}
\usepackage{graphicx}
\usepackage{amsthm,amsmath,amssymb}
\usepackage{mathrsfs}
\usepackage{booktabs}
\usepackage{enumerate}
\usepackage{bbding}
\usepackage{multirow} 
\usepackage{cite}
\usepackage{utfsym}
\usepackage{times}
\usepackage{latexsym}
\usepackage[T1]{fontenc}
\usepackage[utf8]{inputenc}
\usepackage{microtype}
\usepackage{inconsolata}
\usepackage{mathrsfs}
\usepackage{adjustbox}
\title{MEPNet: Medical Entity-Balanced Prompting Network for Brain CT Report Generation}

\author{
    Xiaodan Zhang\textsuperscript{\rm 1},
    Yanzhao Shi\textsuperscript{\rm 1},
    Junzhong Ji\textsuperscript{\rm 1}\thanks{~ Corresponding Authors},
    Chengxin Zheng\textsuperscript{\rm 1},
    Liangqiong Qu\textsuperscript{\rm 2}\footnotemark[1]
}

\affiliations{
    \textsuperscript{\rm 1}College of Computer Science, Beijing University of Technology, Beijing, China\\
    \textsuperscript{\rm 2}School of Computing and Data Science, The University of Hong Kong, Hong Kong, China\\

    zhangxiaodan@bjut.edu.cn,
    yanzhaoshi0927@outlook.com,\\
    jjz01@bjut.edu.cn, 
    chaunxey\_zeta@emails.bjut.edu.cn,
    liangqqu@hku.hk
}

\usepackage{bibentry}

\begin{document}

\maketitle

\begin{abstract}
The automatic generation of brain CT reports has gained widespread attention, given its potential to assist radiologists in diagnosing cranial diseases. 
However, brain CT scans involve extensive medical entities, such as diverse anatomy regions and lesions, exhibiting highly inconsistent spatial patterns in 3D volumetric space.
This leads to biased learning of medical entities in existing methods, resulting in repetitiveness and inaccuracy in generated reports.
To this end, we propose a Medical Entity-balanced Prompting Network (MEPNet), 
which harnesses the large language model (LLM) to fairly interpret various entities for accurate brain CT report generation.
By introducing the \textit{visual embedding} and the \textit{learning status} of medical entities as enriched clues, our method prompts the LLM to balance the learning of diverse entities, thereby enhancing reports with comprehensive findings.
First, to extract visual embedding of entities, we propose Knowledge-driven Joint Attention to explore and distill entity patterns using both explicit and implicit medical knowledge.
Then, a Learning Status Scorer is designed to evaluate the learning of entity visual embeddings, resulting in unique learning status for individual entities.
Finally, these entity visual embeddings and status are elaborately integrated into multi-modal prompts, to guide the text generation of LLM.
This process allows LLM to self-adapt the learning process for biased-fitted entities, thereby covering detailed findings in generated reports.
We conduct experiments on two brain CT report generation benchmarks, showing the effectiveness in clinical accuracy and text coherence.
\end{abstract}

\begin{links}
\link{Code}{https://github.com/YanzhaoShi/MEPNet.}
\end{links}

\section{Introduction}
Radiology reports provide interpretations of detailed pathological findings from medical images, which are crucial for patient treatments.
Despite the significance, for radiologists, hand-crafted writing practices are time-consuming and may cause error outcomes due to subjective factors (e.g. diversion and exhaustion)~\cite{brady2012discrepancy}.
The advent of automated medical report models can produce accurate reports and offer potential benefits for medical examinations, such as reducing the workload shouldered by physicians and saving the limited clinical resources in densely populated countries.

\begin{figure}[t]
\centering  
\includegraphics[width=0.82\linewidth]{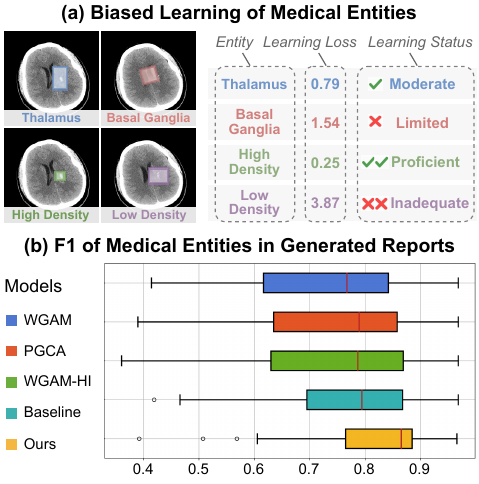}  
\caption{
\textbf{(a)} shows an example of biased entity learning, where different medical entities in brain CT scans exhibit distinct learning losses, indicating biased learning statuses. 
\textbf{(b)} compares F1 scores of different MRG models for covering various medical entities in generated reports. 
Our method shows the narrowest box with a higher median F1 score, indicating our effectiveness in balanced and accurate learning.
}
\label{fig_intro}  
\end{figure}

In recent years, medical report generation (MRG)~\cite{jing2018automatic,yang2021weakly,yang2021radiology, Shi2023Granularity,li2023dynamic} has emerged as a core topic within the medical multi-modal learning.
Different from natural image captioning~\cite{xu2015show,stma}, MRG models focus on reporting tiny pathological conditions from sparse and similar visual information.
This motivates MRG models to learn specific patterns associated with \textit{medical entities}, encompassing entities of \textit{anatomy regions} such as ``basal ganglia'' and entities of \textit{lesions} such as ``high density'' (refer to Figure~\ref{fig_intro}(a)).
To achieve this, task-specific attention mechanisms~\cite{jing2018automatic,wang2018tienet}
are early designed to ground medical semantics with specific visual areas.
Memory mechanism~\cite{chen2022cross,qin2022reinforced}
is then proposed to store cross-modal patterns of diseases for generating accurate content.
Moreover, knowledge graph~\cite{liu2021exploring,li2023dynamic,Shi2023Granularity} is another effective approach for MRG models to grasp prior relations of medical entities.
Recently, as large language models (LLMs) showcase impressive capabilities in language processing, a growing number of studies have started to employ LLMs as decoders for MRG.
Through carefully designing medical prompts~\cite{Bu2024Dynamic,Chen2024Dia} and fine-tuning strategies~\cite{he2024pefomed}, these methods have been proven to be more effective than traditional MRG models.

Despite the burgeoning popularity of LLM-based methods, the focus of most existing models remains on 2D imaging, e.g. chest X-rays.
In contrast, brain CT data encompasses more complex pathological conditions in 3D volumetric space, with a broader range of medical entities~\cite{Shi2023Granularity}.
These medical entities display 
diverse visual distributions, grounded by unique shapes of cerebral organs or contours of intracranial lesions ~\cite{Bhadauria2014Intracranial}.
This visual diversity is vital for radiologists to identify and diagnose.
However, empowering LLMs to effectively capture these diverse visual clues is challenging~\cite{Chen2024Dia, Blankemeier2024Merlin}, which results in a huge learning bias across different entities (see Figure~\ref{fig_intro}(a)).
This bias disperses the language model's focus towards entities that should not be overly emphasized or ignored, leading to the generation of repetitive and inaccurate diagnostic sentences.
Although logit-adjusted classifier~\cite{Jin2024PromptMRG} is recently proposed to mitigate this bias by prompting the language model with well-classified textual labels of entities, it reduces the LLM's sensitivity to interpret visual patterns since it relies too heavily and too early on classified texts.
Thus, how to prompt LLMs to fairly grasp visual patterns across various medical entities remains a critical concern in MRG tasks.

In this paper, we propose a Medical Entity-balanced Prompting Network (MEPNet) to enhance the capability of LLM in balancing the learning of diverse medical entities for high-quality brain CT report generation.
The core idea is to introduce the \textit{visual embedding} and \textit{learning status} for individual medical entities, which are integrated as enriched clues to prompt LLM to fairly understand entities' visual patterns.
Specifically, the learning status is employed to evaluate the quality of entity visual embedding. 
To derive visual embedding for entities, we propose a Knowledge-driven Joint Attention. 
It captures and distills entity patterns from CT scans via explicit medical knowledge (summarized from a knowledge graph developed under expert guidance) and implicit medical knowledge (explored by the model itself).
Next, a Learning Status Scorer is introduced to evaluate the learning statuses of entities, which are then converted to status words, e.g. \textit{limited}, \textit{moderate}, and \textit{proficient}, to represent varying degrees of learning.
Finally, we integrate these entity visual embeddings and related status as entity-balanced prompts, which are elaborately combined with scan visual embeddings and task-specific instructions to form multi-modal prompts, guiding LLM in generating texts.
As evidenced in Figure~\ref{fig_intro}(b), our method achieves a balance in precisely generating diverse medical entities in reports.

Our main contributions can be summarized as:
\begin{enumerate}
\item We propose a novel method that balances the LLM in learning diverse medical entities for accurate brain CT report generation. This is achieved by integrating enriched medical clues, which include visual embeddings and related learning status of entities, to prompt LLM fairly generate findings.
\item We design a Knowledge-driven Joint Attention to exploit high-level visual embeddings of medical entities from 3D volumetric space by using explicit and implicit medical knowledge; We further introduce a Learning Status Scorer to evaluate the quality of entity visual embeddings.
\item We comprehensively evaluate our model on the BCT-CHR and CTRG datasets. The results confirm the effectiveness of our method in generating accurate brain CT reports.
\end{enumerate}

\begin{figure*}[t]
\centering  
\includegraphics[width=0.95\linewidth]{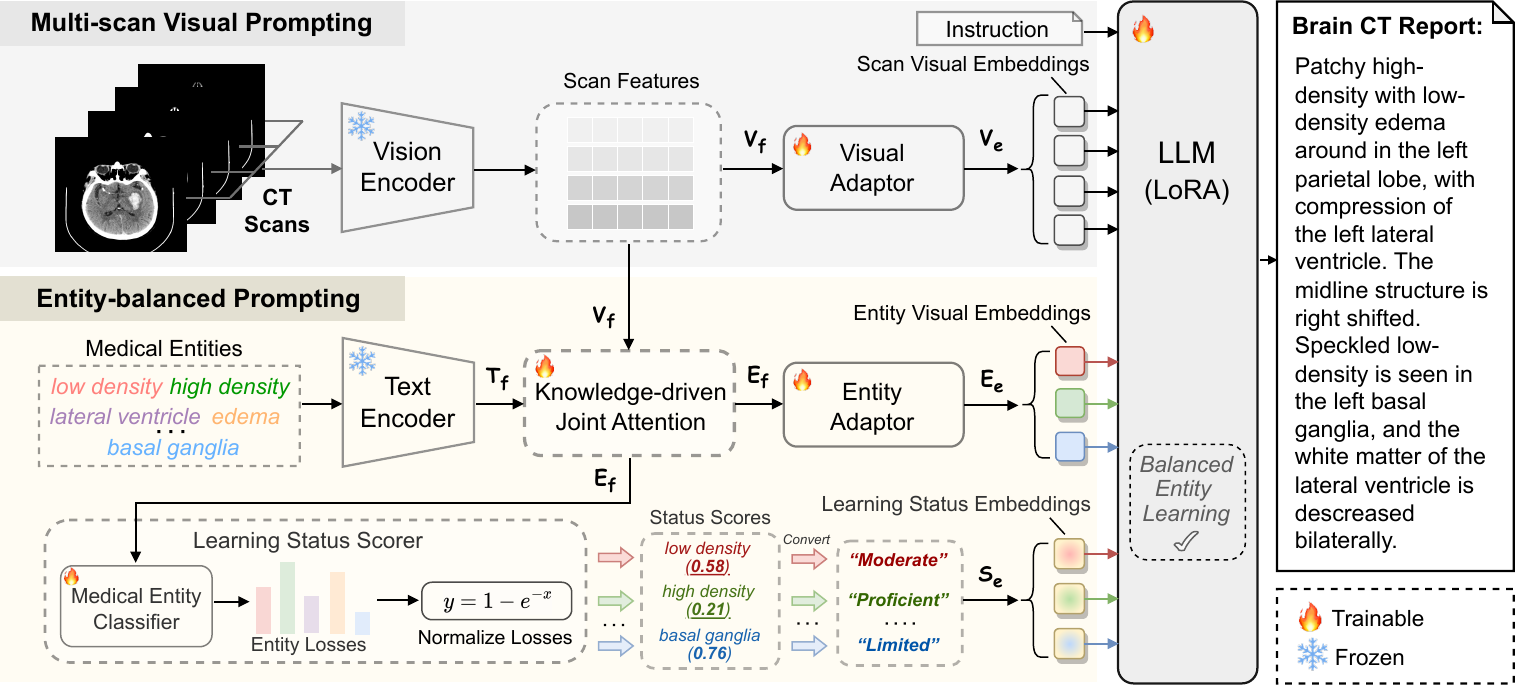}  \caption{
The framework of MEPNet, with two branches: Multi-scan Visual Prompting, which processes visual information of brain CT scans, and Entity-balanced Prompting, which mines entity visual embeddings and corresponding learning status for achieving balanced entity learning within LLM.
These branches collaboratively prompt the LLM to generate diagnostic reports.
}
\label{framework}  
\end{figure*}

\section{Related Works}
With the development of intelligent healthcare, medical report generation (MRG), as an essential yet challenging application, has received notable attention~\cite{jing2018automatic,li2019knowledge,Jia2020bibm,zhang2020when,song2022cross,zhang2023semi,Shen2024GHCL,Chen2024Dia,Zheng2024See}.
To satisfy the accuracy in generating long medical descriptions, existing efforts can be roughly divided into three types.
The first type is enhancing visual features via domain knowledge for producing precise texts.
\citet{jing2018automatic} propose to inject medical concepts into visual embeddings by a co-attention mechanism.
\citet{li2019knowledge} and \citet{zhang2020when} mine disease relations from graph knowledge and use them to enhance visual contexts. 
The second type of MRG is to conduct visual-textual alignment to improve the cross-modal correspondence.
To achieve this, memory mechanism~\cite{chen2022cross,qin2022reinforced,wang2022cross} is designed to map multi-modal features into mutual space for generating accurate sentences grounded by relevant visual patterns.
Contrastive learning~\cite{yan21weakly,Shi2023Granularity, yang2021radiology,li2023dynamic,Shi2024Prior} is also widely applied to model cross-modal representations in an unsupervised manner, showing efficiency in small-scale medical datasets.
The recent advance in LLMs inspires the third type of MRG, which can replace previous complex model designs with unified LLMs while improving task performance~\cite{Zhou2024Large}.
\citet{yan2023style} froze an LLM and leverage in-context learning to generate human-style chest X-ray reports.
Dia-LLaMA~\cite{Chen2024Dia} fine-tunes LLaMA2-7B~\cite{Touvron2023Llama2} by using discriminated disease labels as knowledge prompts.
\citet{Bu2024Dynamic} extract knowledge of pulmonary lesions from patient instances to facilitate X-ray report generation. 
\citet{Liu2024Bootstrapping} further bootstrap LLMs for MRG by instance induction and coarse-to-fine decoding.
Compared with them, we focus on more complex 3D brain CT scans and offer a novel prompting paradigm to represent medical entity clues with informative embeddings and learning status.
This activates the inner power of LLM to balance entity learning and reduce the hallucinations in generating diagnostic sentences.

\section{Methods}
\subsection{3.1 Overall Framework}
An overview of MEPNet is shown in Figure~\ref{framework}. 
Given a patient sample with $N$ brain CT scans $S =\{s_1, ..., s_N\}$, the target is to generate a brain CT report $R =\{r_1, ..., r_M\}$ with $M$ textual words.
We employ LLM as the decoder for report generation, with enriched prompts derived from two distinct branches: The multi-scan visual prompting branch and the entity-balanced prompting branch.

\subsubsection{Multi-scan Visual Prompting}
To obtain visual features from volumetric CT scans, we follow~\citet{zhang2023weakly}
and utilize a finetuned ResNet101 model to extract scan features $V_f =\{v_{f_1}, ..., v_{f_N}\}\in \mathbb{R} ^{N \times P \times d_s}$ for each sample, where $P$ and $d_s$ denote the patch number and feature channels, respectively.
We then employ a visual adaptor to convert $V_f$ into the word embedding space $d_w$ within LLM:
\begin{equation}
    {V_e} = (V_f W_{v_1}^\top+ b_{v_1})W_{v_2}^\top+ b_{v_2},
\end{equation}
where $V_e\in \mathbb{R} ^{N \times d_w}$ denotes the scan visual embedding, $W_{v_1}\in \mathbb{R} ^{(P*d_s) \times d_s}$ and $W_{v_2}\in \mathbb{R} ^{d_s \times d_w}$ are learnable weights, $b_{v_1}$ and $b_{v_2}$ and are learnable biases.

\subsubsection{Entity-Balanced Prompting}
Adding medical information into prompts is proven to enhance the diagnostic capabilities of LLMs~\cite{Jin2024PromptMRG}. 
Unlike previous methods that construct text prompts based on classified categories, we propose to use informative visual embeddings of medical entities as prompts.
This forces the LLM to grasp medical patterns from redundant visual features, rather than just taking shortcuts by rewriting the classified entity labels.
We first collect a set of medical entities $E = \{e_0, e_1, ..., e_k\}$, where $K$ entities $e_{1:k}$ are summarized by \citet{Shi2023Granularity} and $e_0$ denotes a global entity (i.e. $[global]$) to represent overall conditions.
We extract textual features of $k+1$ entities as $T_f\in \mathbb{R} ^{(k+1) \times d_w}$ via LLM's text embedding layer.
Then, $T_f$ is used to capture detailed entity visual features $E_f\in \mathbb{R} ^{(k+1) \times d_h}$ from the whole scan features $V_f$ via Knowledge-driven Joint Attention (KJA) module (see Section 3.2.).
Next, we remove the global entity from $E_f$
and map the remaining feature into LLM's embedding space, resulting in entity visual embeddings $E_e\in \mathbb{R} ^{k \times d_w}$.

Additionally, we introduce a learning status scorer to evaluate the learning status of $k$ entities, which is denoted as $T_s = \{t_{s_1}, ..., t_{s_k}\}\in \mathbb{R} ^{k}$. We convert $T_s$ into status words $S_w = \{s_{w_1}, ..., s_{w_k}\}$, which is embedded as learning status embeddings $S_e$ to depict the learning degree of each entity.

\subsubsection{Model Training}
\begin{figure}[t]
\centering  
\includegraphics[width=0.92\linewidth]{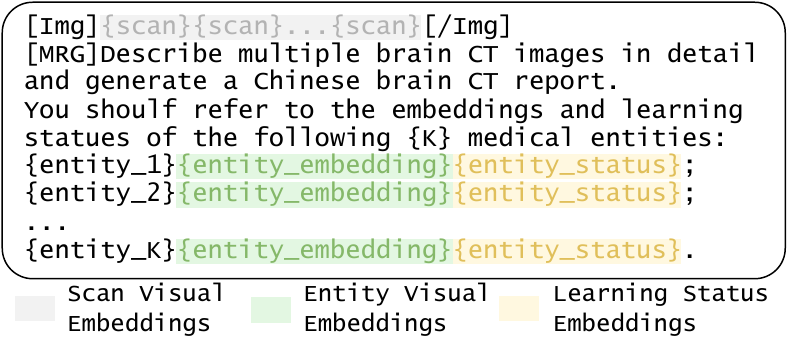}  \caption{
Template of the multi-modal prompt. Different colors are utilized to represent distinct components.
}
\label{prompts}  
\end{figure}

We integrate the scan visual embedding $V_e$, entity visual embedding $E_e$, and learning status embedding $S_e$ into multi-modal prompt $M_P$, which guides a LLM-based decoder
to generate reports. The template of our multi-modal prompt is shown in Figure~\ref{prompts}, where $V_e$, $E_e$, and $S_e$ are denoted as \textit{\{scan\}}, \textit{\{entity\_embedding\}}, and \textit{\{entity\_status\}}, respectively.
The remaining contents are instructions, where \textit{\{entity i\}} is the entity word $e_i$.
We also add special tokens, e.g. \textit{[Img]} and \textit{[/Img]} for visual input identification, and \textit{[MRG]} for task identification.
We finetune LLM by LoRA~\cite{hu2021lora} and use cross-entropy loss for optimization:
\begin{equation}
\mathcal{L}_g = - \sum_{t=1}^M \log P(y_t \mid y_{1:t-1}, M_P; \theta)
\end{equation}
where $P(y_t|*)$ denotes the probability conditioned on multi-modal prompt $M_P$ and the embeddings of previous words \( y_1, y_2, \ldots, y_{t-1} \).
$\theta$ denotes the trainable parameters in LLM.

\subsection{3.2 Knowledge-Driven Joint Attention}
\label{KJA_section}
Given entity textual features $T_f$ and scan features $V_f$, this module aims to capture detailed visual features for entities based on entities' medical relationships.
Unlike directly injecting relations into visual features~\cite{zhang2020when} or textual keywords~\cite{li2023dynamic}, we argue that they may fail to learn multi-modal medical clues.
Instead, we first integrate $T_f$ into $V_f$ by cross attention, and then use knowledge-masked self attention to inject relations.

\subsubsection{Cross Attention}
We project entity textual features $T_f$ into $d_h$ channels to form query vector $Q_c$.
The channel of scan features $V_f$ is also projected to $d_h$ to form key vector $K_c$ and value vector $S_c$.
The cross attention can be formulated as:
\begin{equation}
T^{'}_f = FFN(MHA(Q_c, K_c, S_c),
\end{equation}  
where $T^{'}_f \in \mathbb{R} ^{(k+1) \times d_h}$ primarily depicts the visual features of medical entities. FFN and MHA denote the feed-forward network and multi-head attention, respectively.

\subsubsection{Knowledge Extraction}
The obtained $T^{'}_f$ is still limited to representing complex visual patterns since it overlooks the inherent medical relations of entities that are vital for diagnosis.
To tackle this, we extract entities' relations from both explicit and implicit perspectives as medical knowledge.

\begin{figure}[t]
\centering  
\includegraphics[width=0.94\linewidth]{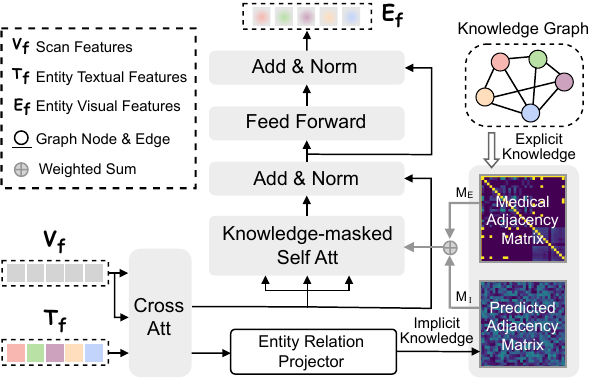}  \caption{
Details of the Knowledge-driven Joint Attention.
}
\label{graph_encoder}  
\end{figure}

Explicit entity relations depict empirical medical knowledge defined by experts.
We acquire this knowledge from brain pathological graph~\cite{Shi2023Granularity}.
This graph divides medical entities $E$ into anatomy regions $E_r \in E$ and lesions $E_l \in E$, with three types of fine-grained edges connecting them: 1)~$R_{t2t}$: edges among anatomy regions; 2) $R_{l2l}$: edges among lesions; 3) $R_{t2l}$: edges between anatomy regions and lesions.
$R_{t2t}$ and $R_{l2l}$ are fixed by medical priors, and $R_{t2l}$ is built based on the specific tissue-lesion relations in ground truth report. 
Note that during model inference, we do not extract $R_{t2l}$ from ground truth report to prevent data leakage. Instead, we substitute it with the statistical information of tissue-lesion relationships derived from the training data.
Based on this graph, given one entity $e_i \in E$ as a query, we can find its relations with other entities, which can be denoted as $R_{i2E} \in \mathbb{R} ^{k+1}$.
The relations associated with the global entities $e_0$ are set to 1.
We extract all entities' relations and concatenate them as medical adjacency matrix $M_E \in \mathbb{R} ^{(k+1) \times (k+1)}$ (see Figure~\ref{graph_encoder}).

Besides, there also exists some implicit yet essential relations among medical entities that haven't been explored by radiologists.
To fully consider these potential entity relations,
we introduce an entity relation projector to predict them automatically, which can be formulated as:
\begin{eqnarray}
    R_{e} &=& W_{r_e} T^{'}_f + b_{r_e}, \\
    M_I &=& Sigmoid(W_{r_p}R_{e} {R^{\top}_{e}} + b_{r_p}),
\end{eqnarray}
where $Sigmoid(.)$ is the sigmoid function. 
$M_I \in \mathbb{R} ^{(k+1) \times (k+1)}$ is the predicted adjacency matrix. 
$W_{r_e}$, $W_{r_p}$ are learnable weight matrixes, $b_{r_e}$, $b_{r_p}$ are learnable biases.

We combine two types of matrixes into the entity adjacency matrix $M_{adj}$, which can be formulated as:
\begin{equation}
\label{adj_matrix}
M_{adj} = \alpha_E M_E + \alpha_I M_I,
\end{equation}  
where $\alpha_E$ and $\alpha_I$ are hypermeters to balance the weights of $M_E$ and $M_I$. 

\subsubsection{Knowledge-Masked Self Attention}
We employ $M_{adj}$ as the attention mask and conduct knowledge-masked self attention to inject medical knowledge into $T^{'}_f$:
\begin{eqnarray}
e_{att}&=&LN(KSA(T^{'}_f, M_{adj}) + T^{'}_f),\\
E_f&=&LN(FFN(e_{att}) + e_{att}),
\end{eqnarray}
where $KSA$ is masked multi-head attention with $M_{adj}$ as mask, $LN$ is the layer norm. $E_f = \{e_{f_0}, ..., e_{f_k}\}\in \mathbb{R} ^{(k+1) \times d_h}$ denotes the entity visual features.
$e_{f_i}$ denotes the visual feature of the i-th entity, which aggregates its neighbor entity features according to relations in $M_{adj}$.

\subsection{3.3 Learning Status Scorer}
\label{pathology_scorer}
To address the biased learning of medical entities, we propose a learning status scorer to evaluate the learning status of each entity at each training step. These learning statuses are used to prompt the LLM to balance the learning of diverse entities.

\subsubsection{Score Generation}
The scorer is based on calculating discriminative loss for each medical entity.
Followed by \citet{Jin2024PromptMRG},
we design a medical entity classifier to obtain discriminative loss. 
The classifier can be formulated as:
\begin{eqnarray}
    E_h &=& W_{e_h} E_f + b_{e_h}, \\
    E_p &=& Sigmoid(Cls(E_h)),
\end{eqnarray}
where $Cls(.)$ denotes a classification head, $E_p = \{e_{p_1}, ..., e_{p_k}\}\in \mathbb{R} ^{k}$ denotes the predicted labels for $k$ entities. 
Then, given a ground truth label $\hat{E}_p = \{\hat{e}_{p_1}, ..., \hat{e}_{p_k}\}\in \{0,1\} ^{k}$, we gather labels of individual entities in a batch level to measure the loss of each entity, e.g., the loss for entity $e_{i}$ can be calculated as:
\begin{equation}
e_{s_i} = - \sum^{b}_{j} W_{e_j} [e^{(j)}_{p_i} \log {\hat{e}}^{(j)}_{p_i} + (1-{e}^{(j)}_{p_i})\log (1-{\hat{e}}^{(j)}_{p_i})],
\end{equation}
where $b$ denotes the batch size, $e^{(j)}_{p_i}$ and ${\hat{e}}^{(j)}_{p_i}$ denote the ground truth label and predicted label of entity $e_{i}$ in the $j-th$ sample, respectively.
Then, we apply the Exponential Saturation Function $y = 1 - e^{-x}$ to normalize the loss to the range of 0 to 1.
\begin{table*}[h]
\centering
\setlength{\tabcolsep}{9.5pt}{
\begin{tabular}{clccccccc}
\toprule
\multirow{2}{*}{\adjustbox{valign=c}{\textbf{Datasets}}} & \multirow{2}{*}{\adjustbox{valign=c}{\textbf{Methods}}} & \multicolumn{4}{c}{\textbf{NLG Metrics}} & \multicolumn{3}{c}{\textbf{CE Metrics}} \\
\cmidrule(r){3-6} \cmidrule(l){7-9}
\rule{2pt}{-4pt}
                         &          & \textbf{B4}            & \textbf{M}             & \textbf{R-L}           & \textbf{C}             & \textbf{F1}            & \textbf{Recall}        & \textbf{Precision}     \\
\midrule
\multirow{8}{*}{\textbf{CTRG-B}}  & Up-Down\cite{anderson2018bottom}${\dagger}$  & 24.4          & 31.6          & 42.5          & 27.3          & \textbf{-}    & \textbf{-}    & \textbf{-}    \\
                         & WCL\cite{yan21weakly}${\dagger}$      & 25.1          & 31.3          & 42.8          & 33.3          & \textbf{-}    & \textbf{-}    & \textbf{-}    \\              
                         & WGAM\cite{yang2021weakly}${\dagger}$     & 25.4          & 32.0          & 42.4          & 31.9          & \textbf{-}    & \textbf{-}    & \textbf{-}    \\
                         & PGCA\cite{Shi2023Granularity}${\dagger}$     & 26.5          & 32.5          & 43.0          & 34.0          & \textbf{-}    & \textbf{-}    & \textbf{-}    \\
                         & WGAM-HI\cite{zhang2023weakly}${\dagger}$  & 26.1          & 31.4          & 43.8          & 33.2          & \textbf{-}    & \textbf{-}    & \textbf{-}    \\
                         &PromptMRG\cite{Jin2024PromptMRG}${\dagger}$  & 26.6          & 31.0          & 47.4          & 50.3          & \textbf{-}    & \textbf{-}    & \textbf{-}    \\
\cmidrule(l){2-9}
                         &HILT\cite{Liu2024Benchmarking}${\dagger}$  & 33.3          & 33.0          & 50.3          & 80.0          & \textbf{-}    & \textbf{-}    & \textbf{-}    \\
                         & MEPNet (Ours)     & \textbf{34.4} & \textbf{34.6} & \textbf{50.4}          & \textbf{86.6} & \textbf{-}    & \textbf{-}    & \textbf{-}    \\
\midrule
\multirow{8}{*}{\textbf{BCT-CHR}} & Up-Down\cite{anderson2018bottom}${\dagger}$  & 13.5          & 27.0          & 35.4          & 20.1          & 47.2          & 59.8          & 49.5          \\
                         & WCL\cite{yan21weakly}${\dagger}$      & 14.2          & 26.9          & 35.3          & 18.8          & 42.4          & 50.2          & 51.0          \\
                         & WGAM\cite{yang2021weakly}${\dagger}$     & 14.5          & 27.8          & 35.1          & 17.3          & 41.6          & 52.6          & \textbf{52.8}          \\
                         & PGCA\cite{Shi2023Granularity}${\dagger}$     & 15.5          & 28.7          & 36.5          & 19.9          & 49.6          & 61.9          & 52.0          \\
                         & WGAM-HI\cite{zhang2023weakly}${\dagger}$  & \textbf{15.6} & 29.0          & 36.5          & 22.2          & 51.4          & 66.8          & 51.1          \\
                         &PromptMRG\cite{Jin2024PromptMRG}${\dagger}$  & 12.9          & 26.8          & 34.0          & 16.1          & 49.6    & 52.9    & 43.3    \\
\cmidrule(l){2-9}
                         &HILT\cite{Liu2024Benchmarking}${\dagger}$  & 12.6          & 27.3          & 33.6         & 15.9          & 43.8    & 52.4    & 39.9    \\
                         & MEPNet (Ours)     & 14.7          & \textbf{29.2} & \textbf{36.8} & \textbf{22.3} & \textbf{52.0} & \textbf{68.9} & 50.6         \\
\bottomrule
\end{tabular}
}
\caption{Performance comparison of our MEPNet with state-of-the-art models on two brain CT report generation datasets. Models are categorized into those utilizing traditional decoders and those employing LLM decoders. Superior results are highlighted in bold. Note that CE metrics are solely evaluated on the BCT-CHR dataset, as the CTRG-Brain dataset lacks a unified evaluation criterion. ${\dagger}$ denotes the re-produced models.
}
\label{comparison}
\end{table*}
\begin{table*}[t]
\centering
\setlength{\tabcolsep}{4.3pt}{
\begin{tabular}{cccccccccccc}
\toprule
\multirow{2}{*}{\adjustbox{valign=c}{\textbf{Method}}} &
  \multicolumn{2}{c}{\textbf{Entity-balanced Prompts}} &
  \multicolumn{2}{c}{\textbf{Entity Adjacency Matrix}} &
  \multicolumn{3}{c}{\textbf{CE Metrics}} &
  \multicolumn{4}{c}{\textbf{NLG Metrics}} \\
\cmidrule(lr){2-3} \cmidrule(lr){4-5} \cmidrule(lr){6-8} \cmidrule(lr){9-12}
 &
  \textbf{\textit{Entity Visual Embed.}} &
  \textbf{\textit{Status Embed.}} &
  \textbf{w/ $M_E$} &
  \textbf{w/ $M_I$} &
  \textbf{F1} &
  \textbf{R} &
  \textbf{P} &
  \textbf{B4} &
  \textbf{M} &
  \textbf{R-L} &
  \textbf{C} \\
\midrule
baseline &   &   &   &   & 33.7 & 43.1 & 46.3          & 10.2 & 24.0 & 32.3 & 12.8 \\
(a)      & $\checkmark$ &   & $\checkmark$ &   & 35.4 & 43.9 & 45.8          & 11.3 & 24.5 & 35.2 & 17.1 \\
(b)      & $\checkmark$ &   &   & $\checkmark$ & 43.9 & 56.1 & 47.9          & 13.1 & 26.5 & 36.7 & 19.0 \\
(c)      & $\checkmark$ &   & $\checkmark$ & $\checkmark$ & 44.0 & 55.9 & \textbf{49.3} & 13.9 & 26.8 & 36.4 & 21.3 \\
(d)      & $\checkmark$ & $\checkmark$ & $\checkmark$ &   & 45.4 & 60.0 & 44.6          & 12.7 & 28.8 & 34.5 & 21.7 \\
(e)      & $\checkmark$ & $\checkmark$ &   & $\checkmark$ & 49.4 & 67.8 & 46.5          & 14.2 & 28.2 & 35.9 & 19.6 \\
Ours     & $\checkmark$ & $\checkmark$ & $\checkmark$ & $\checkmark$ & \textbf{51.4} & \textbf{68.8} & 48.8 & \textbf{14.7} & \textbf{29.2} & \textbf{36.8} & \textbf{22.3} \\
\bottomrule
\end{tabular}
}
\caption{
Ablation studies of the proposed MEPNet on the BCT-CHR dataset, where different settings of medical entity-balanced prompts and knowledge-driven joint attention are evaluated. KJA denotes the integration of knowledge-driven joint attention. \textbf{\textit{$M_E$}} and \textbf{\textit{$M_I$}} denote the medical adjacency matrix and predicted adjacency matrix in KJA, respectively.
}
\label{abaltion_table}
\end{table*}
We use the normalized entity loss as the score of learning status, where a higher score indicates a worse learning of related entity. The scores for all entities are represented as $E_s = \{e_{s_1}, ..., e_{s_k}\}\in \mathbb{R} ^{k}$.

Meanwhile, to ensure a good evaluation of learning status, we optimize the classifier during the training:
\begin{equation}
\mathcal{L}_{d} = - \sum^{b}_{i} W_{d_i} [E^{(i)}_{p} \log \hat{E}^{(i)}_p + (1-{E}^{(i)}_{p})\log (1-\hat{E}^{(i)}_{p})],
\end{equation}
where $E^{(i)}_{p}$ and $\hat{E}^{(i)}_p$ represent the ground truth label and predicted label of sample $i$.

\begin{figure}[h]
\centering  
\includegraphics[width=1.00\linewidth]{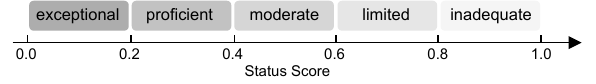}  \caption{
Matching rules of status words and status scores.
}
\label{status_matching}  
\end{figure}

\subsubsection{Status Word Matching}
To let the LLM understand the status scores for balancing entity learning, we convert the generated scores to textual words, as LLM has an inherent sensitivity to interpreting texts.
As shown in Figure~\ref{status_matching}, we set 5 types of status words to match different ranges of scores.
Given scores $E_s\in \mathbb{R} ^{k}$, the converted words can be denoted as $S_w = \{ s_{w_1}, ..., s_{w_k}\}\in \mathbb{R} ^{k}$.
We then employ LLM's text embedding layer to encode $S_w$ into learning status embeddings $S_e \in \mathbb{R} ^{k \times d_w}$.
For status words that are tokenized into multiple tokens by LLM, we average their embeddings to represent the centroid semantics, which reduces computational costs while preserving the initial semantics~\cite{li2023dynamic}.

\subsection{3.4 Training Objective}
Our overall loss contains report generation loss and classification loss, which can be represented as:
\begin{equation}
    \mathcal{L} = \mathcal{L}_g + \lambda\mathcal{L}_d,
\end{equation}
where $\lambda$ is the coefficient to balance the module losses.

\section{Experiments and Results}
\subsection{4.1 Datasets, Metrics and Settings}
\subsubsection{BCT-CHR}
BCT-CHR~\cite{yang2021weakly} is the benchmark for brain CT report generation.
It comprises 49,152 brain CT scans paired with 2,048 Chinese reports. 
Each example includes 24 CT scans along with a report. 
We split the training, testing, and validation sets in a ratio of 7:2:1.
Note that in this paper, we only give English translations of Chinese reports for better understanding.

\subsubsection{CTRG-Brain}
CTRG-Brain~\cite{tang2024work} is an open-source dataset, which contains 6007 samples with cranial abnormalities and includes 160,352 brain CT scans and 6007 Chinese reports. 
We employ brain CT data cleaning pipeline~\cite{zhang2023weakly} to filter redundant scans and ensure each sample contains 24 scans, aligning the data with medical standards.
Since the official split misses a validation set, we align its data split with the BCT-CHR in 7:2:1.

\subsubsection{Metrics}
We use natural language generation (NLG) and clinical evaluation (CE) metrics to evaluate the report generation.
The NLG metrics contains BLEU~\cite{papineni2002bleu}, METEOR~\cite{lavie2007meteor}, ROUGE-L~\cite{lin2004rouge} and CIDEr~\cite{vedantam2015cider}.
For CE metrics, we follow~\citet{Shi2023Granularity} and collect 24 keywords, e.g. ``basal ganglia region", ``cerebral sulcus" and ``stagnant blood", to calculate Precision, Recall, and F1 scores.

\subsubsection{Settings}
Our baseline follows encoder-decoder architecture. We utilize ResNet101~\cite{he2016deep} to process visual features, which is fine-tuned for the brain hemorrhage classification using the CQ500 dataset ~\cite{chilamkurthy2018development}.
For the textual decoder, we utilize LLaMA3-8B~\citep{meta2024introducing}, configured with a 4-bit quantization setting.
LoRA~\citep{hu2021lora} (with a rank of 64) is also adopted for efficient fine-tuning, with 0.34\% trainable parameters in LLM (27.3M).
We use Chinese instructions to prompt LLM, and the template in Figure~\ref{prompts} is translated into English for better clarity.
The feature channels \{$d_h$, $d_s$, $d_w$\}, knowledge weights \{$\alpha_E$, $\alpha_I$\}, and classification loss weight $\lambda$ are set to \{512, 2048, 4096\}, \{0.9, 0.1\}, and 0.1, respectively.
We train the model on two RTX 3090 GPUs via AdamW optimizer with a learning rate of 1e-4 for 10 epochs, and the batch size is set to 4.

\subsection{4.2 Comparision Studies}
We compare our model with advanced brain CT report generation methods WGAM~\cite{yang2021weakly}, PGCA~\cite{Shi2023Granularity}, and WGAM-HI~\cite{zhang2023weakly}.
Besides, we also reproduce some other models initially designed for image captioning (Up-Down~\cite{anderson2018bottom}), chest X-ray MRG (WCL~\cite{yan21weakly}, PromptMRG~\cite{Jin2024PromptMRG}), and chest CT MRG (HILT~\cite{Liu2024Benchmarking}) for comprehensive comparisons.

As shown in Table~\ref{comparison}, our model achieves superior performance on both two datasets, highlighting the effectiveness of our entity-balanced learning strategy. 
In contrast, Up-Down yields inferior results (especially on B4) due to the limited focus on medical information.
WGAM utilizes weakly guided visual attention to learn spatial features, while WGAM-HI enhances it in a multi-scale fashion and achieves higher CE scores.
Compared with traditional decoders, HILT adopts LLM as decoder, which performs even better than competitors originally designed for brain CT MRG on the CTRG-Brain dataset.
Surprisingly, HILT performs worse on a relatively small BCT-CHR dataset. The reason may be that without appropriate prompts to connect the visual encoder and text decoder, standard LLM struggles to efficiently learn from limited medical data.
Although PromptMRG prompts the language model with classified disease labels, it overly relies on the given labels and only gains modest scores.
Notably, our method excels with significant gains in CIDEr, showing the advances in handling the repetitiveness in generated reports.
The SoTA score on Recall also indicates the comprehensive coverage of medical keywords.
In Figure~\ref{fig_intro}(b), we are encouraged to find that our method shows the largest mean and smallest variance for generating diverse entities, indicating the achievement of balanced medical entity learning.
Despite the B4 rises on CTRG-Brain, the score on BCT-CHR remains lower. This may be due to the more flexible writing styles in BCT-CHR reports, as we do not regulate grammar structure or filler words in LLM, leading to sentence rephrasing and reducing multi-gram word matching. Additionally, our model has lower precision, likely because our status embeddings prompt the LLM to generate a broader range of entities, lowering precision but highly boosting recall and F1 scores, which are more beneficial for diagnosis assistance.

\begin{figure*}[h]
\centering  
\includegraphics[width=0.99\linewidth]{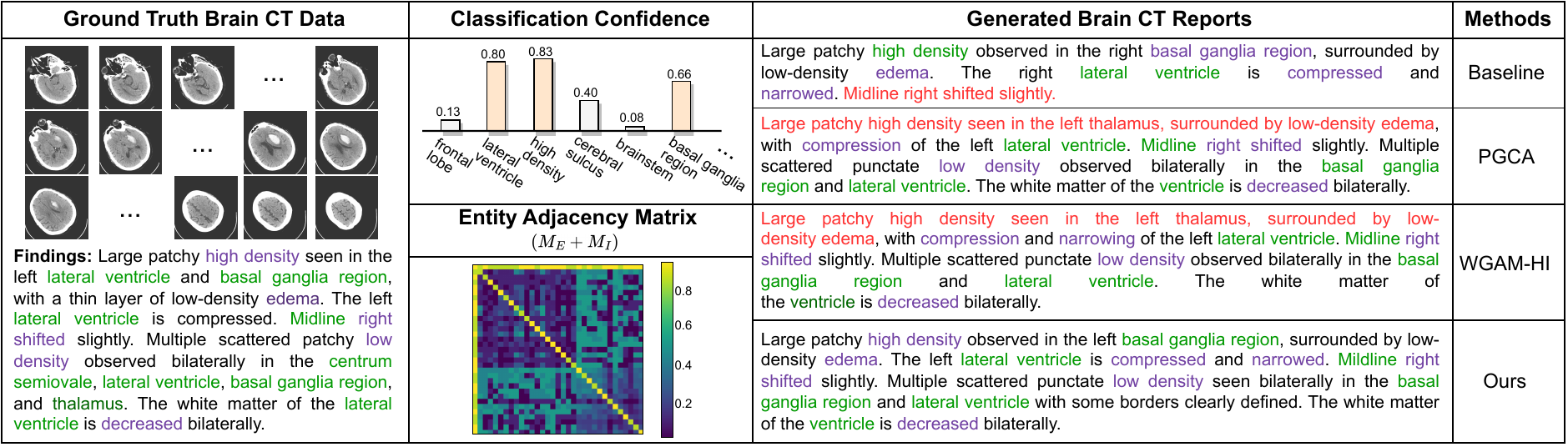}  \caption{
Comparison of brain CT reports generated by different models. Correct medical entities of anatomy region and lesion are marked in green and purple, respectively. Incorrect contents are in red. 
Classification confidence and entity adjacency matrix are also visualized for supplements. 
}
\label{report_visualize}  
\end{figure*}

\subsection{4.3 Ablation Studies}
We conduct ablation studies on components of entity-balanced prompting (i.e. entity visual embedding $E_e$ and status embedding $S_e$) and types of medical knowledge within knowledge-driven joint attention (i.e. $M_E$ and $M_I$).

\begin{table}[ht]
\centering
\setlength{\tabcolsep}{3.2pt}{
\begin{tabular}{cccccccc}
\toprule
\multicolumn{3}{c}{\textbf{Entity Prompts}} &
  \multirow{2}{*}{\textbf{B4}} &
  \multirow{2}{*}{\textbf{M}} &
  \multirow{2}{*}{\textbf{R-L}} &
  \multirow{2}{*}{\textbf{C}} &
  \multirow{2}{*}{\textbf{F1}}\\
\cmidrule(r){1-3}
\textbf{\textit{Embed.}} & \textbf{\textit{Status.}} & \textbf{\textit{Category.}} &               &               &               &               &               \\
\midrule
$\checkmark$ &  &               & 13.9 & 26.8 & 36.4 & 21.3 & 44.0 \\
$\checkmark$ &               & $\checkmark$             & 13.3          & 26.2          & 36.3          & 18.9 & 45.8        \\
$\checkmark$ & $\checkmark$ &               & \textbf{14.7} & \textbf{29.2} & \textbf{36.8} & \textbf{22.3} & \textbf{51.4} \\
$\checkmark$ & $\checkmark$ & $\checkmark$             & 14.2          & 29.1          & 35.4          & 20.6 & 46.6        \\
\bottomrule
\end{tabular}
}
\caption{Comparisons on using different prompts. \textbf{\textit{Embed.}}, \textbf{\textit{Status.}} and \textbf{\textit{Category.}} denote the entity visual embedding, status embedding, and classified entity category, respectively.
}
\label{abaltion_table_cls}
\end{table}

As shown in Table~\ref{abaltion_table}, our baseline performs worse than other settings, which indicates that it is difficult for LLM to directly understand 3D visual features.
Comparing the results of baseline and (a,b,c), we find that using visual embeddings of medical entities substantially enhances the scores, especially when using both the $M_E$ and $M_I$ in KJA (see setting (c)). 
We speculate the reason is that the explicit and implicit medical relations within $M_E$ and $M_I$ effectively enhance entity visual embeddings with medical grounds, thereby guiding LLM to generate precise texts.
Besides, (b) highly improves F1 by 8.5\% over (a), which illustrates the explicit entity relations extracted from the expert knowledge bring more positive effects than predicted relations.
We also observe that (d) and (e) exhibit significant improvements over (a) and (b), respectively, especially in Recall and CIDEr. This indicates the contribution of learning status, which guides LLM to balance the learning of diverse entities and generate precise findings.
Our model's lower precision compared to (c) stems from the reasons discussed in Section 4.2.

To further illustrate the advantages of our entity-balanced prompt, we compare it with prompts that rely on classified categories~\cite{Jin2024PromptMRG, Zhou2024Large}.
We collect the categories (``exist'' or ``not exist'') of each entity, which are predicted by our medical entity classifier. These categories are mapped into textual embeddings (denoted as \textbf{\textit{Category.}}) and added to the prompt.
Comparing the first two lines and the last two lines in Table~\ref{abaltion_table_cls}, it is clear that \textbf{\textit{Category.}} decreases the scores, which is inconsistent with current 2D MRG works.
The reason may be that for more complex 3D data, classified category allows LLMs to take more shortcuts than interpreting visual patterns.
This attributes the upper limit of performance to the classifier, rather than LLM itself.

\subsection{4.4 Qualitative Studies}
In Figure~\ref{report_visualize}, we give examples of brain CT reports to better understand our approach.
As we can see, the baseline generates the worst report with errors and missed findings.
PGCA and WGAM-HI generate more detailed medical content but still have critical errors, such as misdiagnosing the ``\textit{ high density seen in the left thalamus}'' and ``\textit{surrounded by low-density edema}'', which indicates the weakness in capturing diverse medical entities.
In contrast, our model correctly
describes a border range of medical entities, e.g. \textit{basal ganglia}, \textit{lateral ventricles}, and \textit{high density}.
It suggests that, in our approach, LLM is supported by enriched medical clues within entity visual embeddings and learning status, thereby aligning well with ground truth reports.
To further understand the learning process of medical entities, we visualize the confidence score of the medical entity classifier, which manifests aligned results with entities in ground truth reports.
We also visualize the medical entity adjacency matrix (a weighted sum of $M_E$ and $M_I$ calculated by Equation~\ref{adj_matrix}).
It reveals that the model can integrate both explicit and implicit medical knowledge and leverage it for report generation.

\section{Conclusion}
In this paper, we propose to boost brain CT report generation via a medical entity-balanced prompting network, namely MEPNet, which seamlessly enhances the capabilities of the LLM to fairly learn and generate various medical entities.
Experiment results show the effectiveness of our model in comparison with SoTA methods on both the BCT-CHR and CTRG-Brain benchmarks.
We also verify the improvements brought by each prompt component, i.e. entity visual embedding and learning status embedding.
Notably, our approach provides a practical paradigm for activating the capability of LLMs to balance the learning of diverse entities, showing the potential of extending the approach to other medical tasks.

\section*{Acknowledgements}
This work was supported in part by the National Natural Science Foundation of China under Grant 61906007, 62276010, and 62306253, in part by the Guangdong Natural Science Fund-General Programme under Grant 2024A1515010233.

\bibstyle{aaai25}

\end{document}